\documentclass{article} 
\usepackage[preprint]{colm2025_conference}

\usepackage[T1]{fontenc}     
\usepackage[utf8]{inputenc}  
\usepackage{lmodern}         
\usepackage{xcolor}          
\usepackage{microtype}
\usepackage{hyperref}
\usepackage{url}

\usepackage{lineno}
\usepackage{inconsolata}
\usepackage{amsmath}
\usepackage{multirow}
\usepackage{tabularx}
\usepackage{graphicx}
\usepackage{longtable}
\usepackage{booktabs}
\usepackage{setspace}
\usepackage{siunitx}
\usepackage{enumitem}
\usepackage{subcaption}
\usepackage{fancyvrb}
\usepackage{setspace}
\usepackage{float}

\definecolor{darkblue}{rgb}{0, 0, 0.5}
\definecolor{mypink1}{RGB}{248, 206, 204}
\hypersetup{colorlinks=true, citecolor=darkblue, linkcolor=darkblue, urlcolor=darkblue}

\title{Enhancing Faithfulness in Abstractive Summarization \\via Span-Level Fine-Tuning}

\author{
  Sicong Huang \quad Qianqi Yan \quad Shengze Wang \quad Ian Lane \\[4pt]
  University of California, Santa Cruz \\[2pt]
  \texttt{\{shuan213, qyan79, swang488, ialane\}@ucsc.edu}
}

\begin{document}

\ifcolmsubmission
\linenumbers
\fi


\maketitle
\thispagestyle{empty}  
\pagestyle{plain}  

\begin{abstract}
Abstractive summarization using large language models (LLMs) has become an essential tool for condensing information. However, despite their ability to generate fluent summaries, these models sometimes produce unfaithful summaries, introducing hallucinations at the word, phrase, or concept level. 
Existing mitigation strategies, such as post-processing corrections or contrastive learning with synthetically generated negative samples, fail to fully address the diverse errors that can occur in LLM-generated summaries.
In this paper, we investigate fine-tuning strategies to reduce the occurrence of unfaithful spans in generated summaries.
First, we automatically generate summaries for the set of source documents in the training set with a variety of LLMs and then use GPT-4o to annotate any hallucinations it detects at the span-level. 
Leveraging these annotations, we fine-tune LLMs with both hallucination-free summaries and annotated unfaithful spans to enhance model faithfulness.
In this paper, we introduce a new dataset that contains both faithful and unfaithful summaries with span-level labels and we evaluate three techniques to fine-tuning a LLM to improve the faithfulness of the resulting summarization: \textit{gradient ascent}, \textit{unlikelihood training}, and \textit{task vector negation}. 
Experimental results show that all three approaches successfully leverage span-level annotations to improve faithfulness, with \textit{unlikelihood training} being the most effective.
\end{abstract}

\section{Introduction}
Abstractive summarization aims to condense a piece of text into a shorter version by distilling the key information from the source text and rewriting it in a concise manner. Recent advances in large language models (LLMs) such as GPT-4(o) \citep{openai2024gpt4, openai2024gpt4ocard}, Gemini \citep{geminiteam2024geminifamilyhighlycapable}, Llama-3 \citep{grattafiori2024llama3herdmodels}, and Qwen-2.5 \citep{qwen2025qwen25technicalreport} have significantly enhanced the capabilities of summarization systems to produce fluent and coherent summaries.
Additionally, the growing integration of retrieval-augmented generation (RAG)  \citep{NEURIPS2020_6b493230, siriwardhana-etal-2023-improving, zhang2024raft} has underscored the role of summarization as a critical component of modern interactive natural language systems.

However, despite the strong capabilities of LLMs, they still suffer from the problem of hallucination \citep{huang2023survey, Ji_2023, jiang2024survey}, often referred to as unfaithfulness in summarization \citep{maynez-etal-2020-faithfulness, goyal-durrett-2021-annotating, kryscinski-etal-2020-evaluating}. This issue arises when the generated summary contains information that is neither grounded in nor aligned with the source document, limiting the practicality of deploying summarization systems in real applications.
Figure~\ref{fig:examples} shows an example SAMSum \citep{gliwa-etal-2019-samsum} dialogue and its corresponding generated summaries that contains spans of hallucinated text.

A number of approaches have attempted to alleviate unfaithfulness with post-processing. For instance, \citet{dong-etal-2020-multi} and \citet{cao-etal-2020-factual} have suggested methods to edit and correct factual inaccuracies in summaries post-generation. Similarly, \citet{madaan2023selfrefine, akyurek-etal-2023-rl4f} employ the critique-and-refine process that generates critical feedback on the initial summary as a guide for the summarizer to refine the summary. Although effective, the extra post-processing steps required induce high latency and increase the computational demands during inference, restricting their applicable use cases.
Another approach is to learn from negative samples. \citet{cao-wang-2021-cliff, tang-etal-2022-confit, zhang-etal-2023-famesumm, qiu-etal-2024-amrfact} synthetically create negative samples of unfaithful summaries. These samples are derived from reference summaries using strategies that mimic common error types. However, there are three problems with this approach: 
\begin{enumerate}[leftmargin=*]

\item Human reviewers generally prefer LLM summaries over standard reference summaries, even for large and well-established summarization datasets \citep{sottana-etal-2023-evaluation, goyal2023news, liu-etal-2024-learning}, rating them highly across all aspects of evaluation. This preference reveals the poor quality of reference summaries, raising concerns about the effectiveness of fine-tuning models on reference summaries and their perturbed version.

\begin{figure*}[t]
\vspace{-5.5ex}
    \centering
    \small
    \renewcommand{\arraystretch}{1.3}
    \begin{tabular}{|p{0.95\linewidth}|}
        \hline
        \textbf{Source Dialogue:}\\
        \begin{tabular}[c]{@{}p{1.8cm}@{} p{0.85\linewidth}@{}}
        \textit{Pam:} & Hey Robert, you said you could help with Tom's birthday? \\
        \textit{Robert:} & Sure, what do you need? \\
        \textit{Pam:} & I have to go shopping, cook, and clean, and I figured out I don't have time to pick up the balloons. \\
        \textit{Robert:} & From where? \\
        \textit{Pam:} & There's this store in the city centre that sells these awesome floating balloons. \\
        \textit{Robert:} & No problem, just text me the address. \\
        \textit{Pam:} & Bless you! \\
        \textit{Robert:} & ;) \\
        \end{tabular} \\[4pt]
        \hline
        \textbf{Baseline Summary:}\\
        Pam asked Robert for help with Tom's birthday celebration, as she needs to go shopping, cook, and clean, and doesn't have time to pick up floating balloons from a store in the city centre. Robert agreed to help by \colorbox{mypink1}{providing the address of the store}.\\[4pt]
        \hline
        \textbf{Gradient Ascent Summary:}\\
        Pam asked Robert for help with Tom's birthday celebration, including picking up floating balloons from a store in the city centre. Robert agreed to help and requested the store's address.\\[4pt]
        \hline
        \textbf{Unlikelihood Summary:}\\
        Pam asked Robert if he could help with Tom's birthday celebration, specifically asking for his assistance in picking up floating balloons from a store in the city centre. Robert agreed to help and requested the store's address.\\[4pt]
        \hline
        \textbf{Task Vector Negation Summary:}\\
        Pam asked Robert for help with Tom's birthday celebration, including shopping, cooking, and cleaning. Robert agreed to help and \colorbox{mypink1}{Pam provided the address of a store in the city centre} \colorbox{mypink1}{where she needed him to pick up floating balloons}.\\[4pt]
        \hline
    \end{tabular}
    \caption{An example from the SAMSum dataset, showing a dialogue and corresponding summaries generated from models fine-tuned using four different approaches. Unfaithful spans are \colorbox{mypink1}{highlighted}.}
    \label{fig:examples}
\vspace{-3.5ex}
\end{figure*}

\item Synthetic negative samples generated from common approaches often fail to replicate the actual errors observed in model-generated summaries. Furthermore, the error distributions in generated summaries can vary significantly across different domains, rendering synthetic negative sample generation approaches insufficient to cover the wide variety of error types \citep{goyal-durrett-2021-annotating}.

\item Contrastive learning approaches \citep{cao-wang-2021-cliff, tang-etal-2022-confit, zhang-etal-2023-famesumm} lack utilization of detailed, span-level information that could potentially improve summary faithfulness more effectively \citep{goyal-durrett-2021-annotating}. In cases of unfaithfulness within LLM-generated summaries, typically only a few specific text spans are unfaithful. Only these problematic spans should be specifically targeted as negative examples during model training.
\end{enumerate}

\vspace{-1mm}
To address these challenges, we propose annotating span-level hallucinations in LLM-generated summaries and then leveraging this fine-grained information to update the model. Our contributions are threefold: (1) we construct a novel dataset containing LLM-generated summaries, labeled at the span-level for faithfulness; (2) we evaluate three fine-tuning methods, namely gradient ascent, unlikelihood training and task vector negation to utilize the span-level unfaithful information to enhance LLM summarization faithfulness. We show that all three approaches improve faithfulness, with the latter two being more effective; and (3) we analyze how each method is affected by the weight given to negative samples ($\epsilon$), revealing that unlikelihood training is the most stable of the three methods.




\vspace{-1.5ex}
\section{Related work}
\vspace{-1.5ex}
\subsection{Improving summarization faithfulness}
\vspace{-1.5ex}
A number of prior approaches to improving faithfulness focus on post-processing. \citet{dong-etal-2020-multi} uses QA span fact correction models to revise entities in generated summaries to boost factual consistency. Similarly, \citet{cao-etal-2020-factual} corrects factual errors in generated summaries by training a corrector model on artificially created error data.

Other prior works leveraged synthetic negative sample summaries to improve faithfulness.
\citet{cao-wang-2021-cliff} surveyed common errors that summarization models tend to make and designed strategies for constructing negative samples (e.g., entity swap, mask and regenerate) that corrupt the reference summaries. They then used contrastive learning to better discriminate between positive and negative examples, improving the representation and faithfulness during generation.
Similarly, \citet{tang-etal-2022-confit} designed a linguistically informed taxonomy of factual errors for dialogue summaries and created synthetic negative samples based on the taxonomy before applying contrastive learning to improve faithfulness.
\citet{laban-etal-2023-summedits} proposed a cost effective protocol to create more natural sounding and manually verified synthetic negative samples, which can be used as training data for contrastive learning. 

\citet{chen-etal-2021-improving} proposes to generate multiple contrastive candidate summaries featuring different entities from the source document,
subsequently employing a discriminative model trained to differentiate between faithful summaries and synthetic negative ones, effectively ranking the candidates.
\citet{goyal-durrett-2021-annotating} tried to tackle the problem from the perspective of training data. They demonstrated an approach to improve faithfulness by identifying unsupported facts in the training summaries and ignore the corresponding tokens during training.
\citet{goyal-etal-2022-training} took a closer look at the training dynamics and found that longer training on noisy datasets contributes to factual inconsistency. They show that using token sub-sampling to dynamically modify the loss computation during training, down weighting high-loss tokens, can substantially improve factual consistency.

Some more recent methods have focused on the critique-and-refine approach. \citet{akyurek-etal-2023-rl4f} implemented a reinforcement learning framework where a critic model provides feedback on generated summaries. The summarizer then refines its output based on this feedback, with the summarization task metric serving as a reward for the critic model. This process enables the critic to guide the summarizer towards improving specific performance metrics, fostering a cycle of continuous improvement.
\vspace{-1.5ex}
\subsection{Span-level hallucination labeling}
\vspace{-1.5ex}
There has been a large amount of work exploring evaluation metrics for summarization faithfulness \citep{Zhang2020BERTScore:, yuan2021bartscore, liu-etal-2023-g, zha-etal-2023-alignscore}. However, the study of span-level hallucination labeling remains relatively underexplored.
\citet{zhou-etal-2021-detecting} proposed a task to predict token-level hallucinations in summarization and machine translation and introduced methods to create and fine-tune on synthetic data to solve the task.
\citet{goyal-durrett-2021-annotating} presented an approach to label hallucinations in summaries at the dependency arc level, providing more fine-grained information.
Though not specific to summarization, \citet{liu-etal-2022-token} introduced a token-level reference-free hallucination detection benchmark and created multiple baselines.
\citet{niu-etal-2024-ragtruth} introduced RAGTruth, a large-scale hallucination corpus with span-level annotations designed for retrieval-augmented generation (RAG). While primarily focused on hallucination detection in RAG-based applications, the dataset includes extensive word-level annotations across various tasks, including summarization. Their experiments demonstrate that fine-tuning on high-quality annotated data can significantly improve hallucination detection over prompt-based and self-verification methods.

\vspace{-1.5ex}
\section{Methods}
\vspace{-1.5ex}
\subsection{Problem formulation}
\begin{figure*}[ht]
    \centering
    \includegraphics[width=\textwidth]{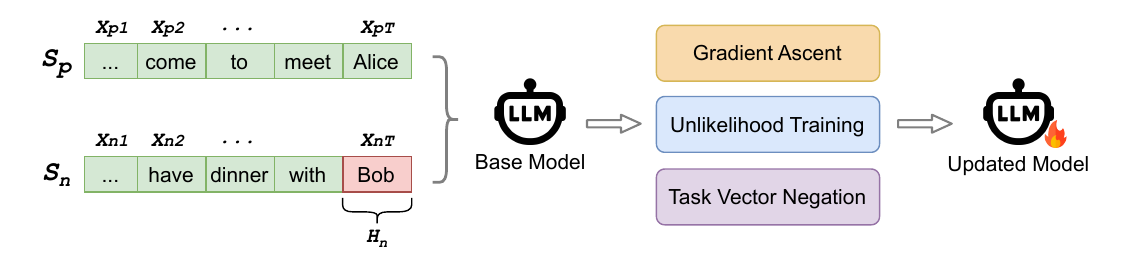}
    \caption{\textbf{Model update:} a base model is updated using both the faithful positive example summaries and the unfaithful negative example summaries with hallucination spans using one of three approaches we compare in this paper (1) Gradient Ascent, (2) Unlikelihood Training or (3) Task Vector Negation.}
    \label{fig:model_update}
\end{figure*}

\vspace{-1.5ex}
As illustrated in Figure \ref{fig:model_update}, training data comprises of both positive and negative samples of document-summary pairs. For each positive sample, the document $D$ has a corresponding positive summary $S_p$ where $S_p$ = $(x_{p1}, x_{p2}, ..., x_{pT})$. Similarly, each negative sample includes a document $D$ paired with a negative summary $S_n$ where $S_n$ = $(x_{n1}, x_{n2}, ..., x_{nT})$. Here, $x$ denotes a token in the summary, and $T$ denotes the number of tokens in the summary. Note that not all $x_n$ tokens are considered hallucinated. Therefore, we use $H_n$ to denote the set of hallucinated tokens in a negative sample summary $S_n$ and $H_n \subseteq S_n$.

\begin{figure*}[ht]
    \centering
    \includegraphics[width=\textwidth]{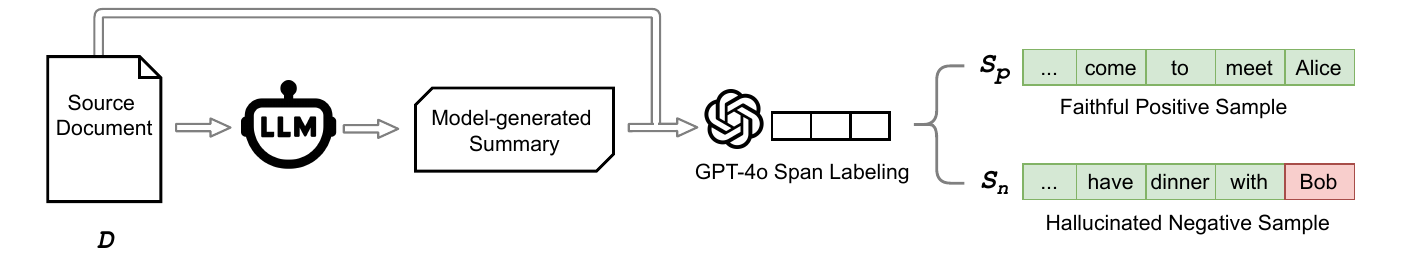}
    \caption{\textbf{Training data construction:} summaries of the source documents used for model training are generated using an LLM. Spans of text in the generated summaries that are unfaithful to the source document are automatically labeled using GPT-4o (using the prompt in Appendix A.3). Summaries that have no unfaithful spans labeled in their output are treated as positive training samples, and summaries that contain unfaithful spans are treated as negative training samples.}
    \label{fig:data_construction}
\vspace{-3ex}
\end{figure*}

\subsection{Fine-tuning methods to improve summarization faithfulness}
\label{sec:methods_studied}
\vspace{-1ex}

We compare three methods for model fine-tuning that can take advantage of both positive faithful summaries and negative unfaithful summaries with span-level hallucination labels. To manage the influence of positive and negative samples, we introduce a hyperparameter $\epsilon \in [0, 1]$ to distribute the weights assigned to each type of summary. Specifically, $\epsilon$ specifies the weight of negative samples, and its complement $1 - \epsilon$ specifies the weight of positive samples.

\vspace{-3ex}
\paragraph{Gradient ascent}
Recent research by \citet{yao2024large} has demonstrated that unlearning undesirable behaviors in LLMs through gradient ascent is an effective alignment technique. This method has been used to steer LLM outputs away from harmful content, copyright infringements, and to reduce hallucinations. 
In our study, we apply gradient ascent to decrease hallucinated content in summarization by minimizing the probability of generating unfaithful tokens. This is achieved by reversing the sign of the cross-entropy loss. We define the gradient ascent loss as follows:
\[
L_{ga} = 
\begin{cases}
    -(1 - \epsilon) \sum \limits_{x_p \in S_p} \log p_\theta (x_p |.) & \text{if } S_p \\
    \epsilon \sum \limits_{x_n \in H_n} \log p_\theta (x_n |.) & \text{if } S_n
\end{cases}
\]
Here, $p_\theta$ represents the model parameterized by $\theta$.
The loss computation varies depending on the sample type: for a positive sample $S_p$, the negative log-likelihood (NLL) loss is calculated with a weight of $(1 - \epsilon)$; for a negative sample $S_n$, the loss is computed as the negative NLL, multiplied by the weight $\epsilon$.

\paragraph{Unlikelihood training}
Unlikelihood training, initially introduced by \citet{Welleck2020Neural}, was designed to mitigate common issues such as dull and repetitive outputs from language models while maintaining perplexity. 
Subsequent studies \citep{li-etal-2020-dont} have demonstrated its effectiveness in addressing problems like excessive copying from context, frequent word overuse, and logical inconsistencies in generated texts. The adaptability of this method allows for its application across various tasks, particularly in reducing undesirable outputs.
In this method, the model is trained to assign lower probabilities to unwanted generations by maximizing the complement of the probability of generating such tokens. For our specific application in summarization, we focus on minimizing the generation of hallucinated tokens. We define the unlikelihood loss as follows:
\[
L_{ul} = 
\begin{cases}
    -(1 - \epsilon) \sum \limits_{x_p \in S_p} \log p_\theta (x_p |.) & \text{if } S_p \\
    -\epsilon \sum \limits_{x_n \in H_n} \log (1 - p_\theta (x_n |.)) & \text{if } S_n
\end{cases}
\]
This configuration computes the negative log-likelihood (NLL) for positive samples, weighted by $(1 - \epsilon)$, and for negative samples, it calculates the NLL of the complement probability $(1 - p_\theta (x_n |.))$ weighted by $\epsilon$. This approach ensures a strategic penalization of unfaithful tokens, thus aiming to enhance the overall faithfulness of the generated summaries.
\vspace{-1ex}
\paragraph{Task vector negation}
Task vector arithmetic offers a straightforward method for modifying a pre-trained model to promote desired behaviors, as outlined by \citet{ilharco2023editing}. This technique involves calculating a task vector, $\tau_t$, for a specific task $t$ by subtracting the weights of the base pre-trained model ($\theta_{pre}$) from the weights of a model that has been fine-tuned on that task ($\theta_{ft}$). Formally, the task vector is defined as $\tau_t = \theta_{ft} - \theta_{pre}$.
Building on this concept, \citet{ilharco2023editing} demonstrated that negating a task vector fine-tuned on toxic language can effectively diminish the generation of toxic outputs. Inspired by this, we apply a similar approach to address the generation of unfaithful summaries. We define the resulting model's weights as:
\vspace{-0.5ex}
\[\theta_{res} = \theta_{pre} + (1 - \epsilon)\tau_{pos} - \epsilon \tau_{neg}\]
$\tau_{pos}$ represents the task vector derived from fine-tuning on positive faithful summaries, while $\tau_{neg}$ is obtained from fine-tuning on negative hallucinated summaries. This method allows us to manipulate the model composition to reduce the production of unfaithful summaries by strategically negating the influence of the undesired traits encoded in $\tau_{neg}$.

\vspace{-1ex}
\section{Data}
\vspace{-1ex}
Recent studies \citep{sottana-etal-2023-evaluation, goyal2023news} have show that even open-source LLMs outperform the reference summaries in well-established datasets, which are often rated poorly on \textit{relevance}, \textit{fluency}, \textit{coherence}, and \textit{consistency}. Additionally, prior methods that generate synthetic negative samples based on common error types \citep{cao-wang-2021-cliff, tang-etal-2022-confit, zhang-etal-2023-famesumm} fail to capture real hallucination patterns in LLM outputs \citep{goyal-durrett-2021-annotating}. Error distributions also vary across domains, making synthetic samples insufficient. To address these issues, we construct a dataset of real LLM-generated summaries that are subsequently annotated for hallucinations at the span-level.
\vspace{-1ex}
\subsection{Dataset construction}
\label{sec:data-construction}
\vspace{-1ex}
The dataset construction process is illustrated in Figure \ref{fig:data_construction}. We construct the training and test set from CNNDM \citep{nallapati-etal-2016-abstractive}, SAMSum \citep{gliwa-etal-2019-samsum}, and XSum \citep{narayan-etal-2018-dont} covering both news and dialogue, two of the most studied domains of summarization research.
Summaries of their source documents were generated using four LLMs, \texttt{Llama3.2-1b}, \texttt{SmolLM2-1.7b}, \texttt{OLMo2-7b}, and \texttt{Llama3.1-8b} \citep{grattafiori2024llama3herdmodels, allal2025smollm2smolgoesbig, olmo20252olmo2furious}, two models for each size class. We set top\_p = 0.7 and temperature = 0.01 to minimize randomness. The prompts we used to generate these summaries are provided in Appendix~\ref{app:prompts}).

\citet{sottana-etal-2023-evaluation} shows that top-tier LLMs correlate highly with human judgments when acting as a reviewer. Expanding upon this prior work we use GPT-4o \citep{openai2024gpt4ocard} to label LLM-generated summaries to identify "spans of text that are inconsistent with the source document." The prompt we use to label these spans is provided in \ref{app:gpt-prompt}).
When a summary has no labeled span in it's output, it is considered a positive example of a summary; if a span has been labeled as "inconsistent with the source document" the summary is treated as a negative example.
\vspace{-0.5ex}
\subsection{Dataset statistics}
\vspace{-1.5ex}

Our dataset consists of 111,897 training samples and 2,819 unlabeled test samples (1,000 CNNDM, 819 SAMSum and 1,000 XSum). Table \ref{tab:dataset} presents the training set's distribution of positive and negative samples across datasets and models. A clear trend emerges: smaller models are more prone to generating unfaithful summaries. Specifically, Llama3.2-1b produces unfaithful summaries in 64\% of CNNDM, 83\% of SAMSum, and 70\% of XSum instances, whereas Llama3.1-8b exhibits significantly lower rate of unfaithfulness—17\% for CNNDM, 30\% for SAMSum, and 25\% for XSum.

Additionally, Table \ref{tab:dataset} shows that smaller models not only generate more unfaithful summaries overall but also produce a higher proportion of hallucinated tokens, as measured by the Hallucinated Token Ratio (HTR). In SAMSum, the dataset where models tend to be the most unfaithful, Llama3.2-1b generates, on average, 42.6\% hallucinated tokens in unfaithful summaries, compared to 22.9\% for Llama3.1-8b. This pattern remains consistent across all three datasets, reinforcing the correlation between model size and summary faithfulness.



\begin{table}[h]
    \centering
    \renewcommand{\arraystretch}{1.2} 
    \resizebox{\textwidth}{!}{%
    \begin{tabular}{l|rrr|rrr|rrr|r}
        \toprule
        \multirow{2}{*}{\textbf{Model}} & \multicolumn{3}{c|}{\textbf{CNNDM}} & \multicolumn{3}{c|}{\textbf{SAMSum}} & \multicolumn{3}{c|}{\textbf{XSum}} & \multirow{2}{*}{\textbf{Total}} \\
        \cmidrule(lr){2-4} \cmidrule(lr){5-7} \cmidrule(lr){8-10}
        & \textbf{Pos} & \textbf{Neg} & \textbf{HTR} & \textbf{Pos} & \textbf{Neg} & \textbf{HTR} & \textbf{Pos} & \textbf{Neg} & \textbf{HTR} & \\
        \midrule
        Llama3.2-1b  & 2751  & 4814  & 0.141 & 1276  & 6384  & 0.426 & 2470  & 5796  & 0.198 & 23491 \\
        SmolLM2-1.7b & 5050  & 6067  & 0.106 & 5492  & 5981  & 0.300 & 2761  & 3576  & 0.155 & 28927 \\
        OLMo2-7b     & 11649 & 3328  & 0.098 & 2889  & 1264  & 0.255 & 5250  & 3372  & 0.142 & 27752 \\
        Llama3.1-8b  & 12287 & 2570  & 0.080 & 4707  & 2047  & 0.229 & 7637  & 2479  & 0.116 & 31727 \\
        \midrule
        \textbf{Total} & 31737 & 16779 &  & 14364 & 15676 &  & 18118 & 15223 &  & 111897 \\
        \bottomrule
    \end{tabular}%
    }
    \caption{The training set's distribution of faithful (\textbf{pos}) and unfaithful (\textbf{neg}) summaries across datasets and models, along with Hallucinated Token Ratio (\textbf{HTR}), the proportion of hallucinated tokens within unfaithful summaries.}
    \label{tab:dataset}
\end{table}

\vspace{-1.5ex}
\section{Experimental setup}
\vspace{-0.5ex}
\subsection{Model and training}
\vspace{-1.5ex}
We perform experiments on all 4 models used for generating the training set (i.e. \texttt{Llama3.2-1b}, \texttt{SmolLM2-1.7b}, \texttt{OLMo2-7b}, and \texttt{Llama3.1-8b}).
For efficient fine-tuning, we adopt Low-Rank Adaptation (LoRA) \citep{hu2022lora} and apply low-rank update matrices to all linear modules, 
with rank $r = 128$, $\alpha = 256$, and dropout probability of LoRA layer = 0.05. With these settings, trainable parameters ranges from 4\% (Llama3.1-8b) to 7.8\% (SmolLM2-1.7b).
The training processes use an AdamW optimizer with learning rate = 5e-5 and linear learning rate scheduler with warm up ratio = 0.01. We set the batch size to 16 and train each model for one epoch on the training set for all experiments.

The baseline is supervised fine-tuning only on the positive portion of training data using regular cross-entropy loss, with no information from negative samples.
\vspace{-1.5ex}
\subsection{Evaluation}
\vspace{-1.5ex}


We evaluate summarization faithfulness using three reference-free automatic metrics: G-Eval, AlignScore, and BARTScore.


\textbf{G-Eval (GE)} \citep{liu-etal-2023-g} is a GPT-based evaluation metric that employs Chain-of-Thought (CoT) prompting \citep{wei2022chain} to generate intermediate reasoning steps before assigning a final score. We use GPT-4o as the underlying model and report its consistency score, which measures the factual alignment between the summary and the source document \citep{10.1162/tacl_a_00373}. 


\textbf{AlignScore (AS)} \citep{zha-etal-2023-alignscore} formulates faithfulness evaluation as a text-to-text information alignment problem. It estimates factual consistency by assessing how well the generated summary preserves key information from the source document. AlignScore is trained to generalize across diverse domains and has demonstrated strong performance in factuality alignment evaluation tasks.


\textbf{BARTScore (BS)} \citep{yuan2021bartscore} models evaluation as a sequence-to-sequence generation task. It measures faithfulness by computing the log probability of generating the source document from the summary using a BART model. A higher BARTScore indicates better faithfulness.


\section{Results and discussion}
The evaluation results for the supervised fine-tuning (SFT) baselines, in which the models have only been updated using faithful summaries ($S_p$) are shown in Table \ref{tab:baseline_model_results} and \ref{tab:baseline_dataset_results}, and the results of the three model update methods explored in this paper are shown in Table \ref{tab:samsum_result} and \ref{tab:llama-8b-result}. Additional results are provided in Appendix~\ref{app:more_results}.


\begin{table}[h]
\centering
\begin{minipage}{0.48\linewidth}
    \centering
    \begin{tabular}{@{}lccc@{}}
        \toprule
        \textbf{Model} & \textbf{GE} & \textbf{AS} & \textbf{BS} \\
        \midrule
        Llama3.2-1b & 4.035 & 0.868 & -3.091 \\
        SmolLM2-1.7b & 4.170 & 0.871 & -3.150 \\
        OLMo2-7b & 4.528 & 0.902 & -3.200 \\
        Llama3.1-8b & 4.581 & 0.900 & -3.143 \\
        \bottomrule
    \end{tabular}
    \caption{SFT baseline results of four models on SAMSum.}
    \label{tab:baseline_model_results}
\end{minipage}
\hfill
\begin{minipage}{0.48\linewidth}
    \centering
    \begin{tabular}{@{}lccc@{}}
        \toprule
        \textbf{Dataset} & \textbf{GE} & \textbf{AS} & \textbf{BS} \\
        \midrule
        CNNDM & 4.733 & 0.900 & -1.906 \\
        SAMSum & 4.581 & 0.900 & -3.143 \\
        XSum & 4.585 & 0.906 & -1.836 \\
        \bottomrule
    \end{tabular}
    \caption{SFT baseline results of Llama3.1-8b on three datasets.}
    \label{tab:baseline_dataset_results}
\end{minipage}
\end{table}

\subsection{Main results}
\label{sec:main-results}


\begin{table}[h!]
\resizebox{\textwidth}{!}{%
\begin{tabular}{lp{2.5cm}cccccccccc}
\toprule
\multicolumn{1}{c}{} & \multirow{2}{*}{\raggedright \textbf{Model}}
& \multicolumn{3}{c}{\textbf{Gradient Ascent}}
& \multicolumn{3}{c}{\textbf{Unlikelihood}}  
& \multicolumn{3}{c}{\textbf{Task Vector}} \\ 
\cmidrule(l){3-5} 
\cmidrule(l){6-8}
\cmidrule(l){9-11}
\multicolumn{1}{c}{} & \multicolumn{1}{c}{} & \textbf{GE} & \textbf{AS} & \textbf{BS} & \textbf{GE} & \textbf{AS} & \textbf{BS} & \textbf{GE} & \textbf{AS} & \textbf{BS}\\
\midrule
& \raggedright Llama3.2-1b  
  & 4.094\(\uparrow\)  & 0.873\(\uparrow\) & -3.099  
  & \textbf{4.166}\(\uparrow\) & \textbf{0.896}\(\uparrow\) & -3.109  
  & 4.010 & 0.878\(\uparrow\) & -3.160  \\

& \raggedright SmolLM2-1.7b  
  & 4.176\(\uparrow\) & 0.876\(\uparrow\) & -3.133\(\uparrow\)  
  & \textbf{4.342}\(\uparrow\) & \textbf{0.896}\(\uparrow\) & -3.139\(\uparrow\)  
  & 4.231\(\uparrow\) & 0.890\(\uparrow\) & -3.162  \\

& \raggedright OLMo2-7b  
  & 4.546\(\uparrow\) & 0.906\(\uparrow\) & -3.153\(\uparrow\)  
  & 4.675\(\uparrow\) & 0.917\(\uparrow\) & -3.164  
  & \textbf{4.733}\(\uparrow\) & \textbf{0.918}\(\uparrow\) & -3.395  \\

& \raggedright Llama3.1-8b  
  & 4.641\(\uparrow\) & 0.918\(\uparrow\) & -3.158  
  & 4.749\(\uparrow\) & 0.918\(\uparrow\) &  -3.198  
  & \textbf{4.760}\(\uparrow\) & \textbf{0.928}\(\uparrow\) & -3.213  \\

\bottomrule
\end{tabular}
}
\caption{Results on the SAMSum dataset for the three studied methods on four models across G-Eval (GE), AlignScore (AS), and BARTScore (BS) evaluation metrics. Gradient ascent $\epsilon = 0.01$, unlikelihood $\epsilon = 0.5$, and task vector $\epsilon = 0.3$. The best result (in terms of G-Eval and AlignScore) for each model is \textbf{in-bold}. Scores higher than the baseline are marked with \(\uparrow\).}
\label{tab:samsum_result}
\end{table}


\begin{table}[h!]
\resizebox{\textwidth}{!}{%
\begin{tabular}{lp{2.5cm}cccccccccc}
\toprule
\multicolumn{1}{c}{} & \multirow{2}{*}{\raggedright \textbf{Dataset}}
& \multicolumn{3}{c}{\textbf{Gradient Ascent}}
& \multicolumn{3}{c}{\textbf{Unlikelihood}}  
& \multicolumn{3}{c}{\textbf{Task Vector}} \\ 
\cmidrule(l){3-5} 
\cmidrule(l){6-8}
\cmidrule(l){9-11}
\multicolumn{1}{c}{} & \multicolumn{1}{c}{} & \textbf{GE} & \textbf{AS} & \textbf{BS} & \textbf{GE} & \textbf{AS} & \textbf{BS} & \textbf{GE} & \textbf{AS} & \textbf{BS}\\
\midrule
& \raggedright CNNDM  
  & 4.737\(\uparrow\) & 0.901\(\uparrow\) & -1.907  
  & \textbf{4.796}\(\uparrow\) & 0.915\(\uparrow\) & -1.875\(\uparrow\)  
  & 4.785\(\uparrow\) & \textbf{0.924}\(\uparrow\) & -1.930  \\

& \raggedright SAMSum  
  & 4.641\(\uparrow\) & 0.918\(\uparrow\) & -3.158  
  & 4.749\(\uparrow\) & 0.918\(\uparrow\) & -3.198  
  & \textbf{4.760}\(\uparrow\) & \textbf{0.928}\(\uparrow\) & -3.213  \\

& \raggedright XSum  
  & 4.632\(\uparrow\) & 0.909\(\uparrow\) & -1.836  
  & \textbf{4.702}\(\uparrow\) & 0.919\(\uparrow\) & -1.808\(\uparrow\)  
  & 4.686\(\uparrow\) & \textbf{0.935}\(\uparrow\) & -1.875  \\

\bottomrule
\end{tabular}
}
\caption{Llama3.1-8b results of 3 studied methods on 3 datasets. Other settings are the same as Table \ref{tab:samsum_result}. The best result (according to G-Eval and AlignScore) for each dataset is \textbf{in-bold}. Scores higher than the baseline are marked with \(\uparrow\).}
\label{tab:llama-8b-result}
\end{table}




First, in terms of G-Eval and AlignScore, regardless of the model, all studied methods show improvements from the baseline. Comparing Table \ref{tab:baseline_model_results} \& \ref{tab:samsum_result}, unlikelihood training yields greater improvements on smaller models (1b and 1.7b) and task vector negation show greater improvements on larger models (7b and 8b). 
The effectiveness of the studied methods is evident in Table \ref{tab:baseline_dataset_results} \& \ref{tab:llama-8b-result} as well, observing improved G-Eval and AlignScore on all three datasets.

Second, in terms of G-Eval and AlignScore, unlikelihood training and task vector negation are more effective than gradient ascent. Gradient ascent, obtains higher scores than the baseline across all models and datasets, but often only outperforms the baseline by a slight margin. This effect is further corroborated by Figure \ref{fig:epsilon} that unlikelihood training and task vector negation show more consistent improvements than gradient ascent. 

BARTScore, however, paints an inconsistent and often contradictory picture: the BARTScore of most results are lower than the baseline, while showing some improvements sporadically. Thus making it hard to draw conclusions from. We suspect the training objectives of the three methods we investigate render BARTScore unsuitable for evaluating faithfulness in this setting. We discuss this further in section \ref{sec:discussion}.

\subsection{Results of varying $\epsilon$}
\label{sec:effect_of_epsilon}
\begin{figure*}[ht]
    \centering
    \includegraphics[width=\textwidth]{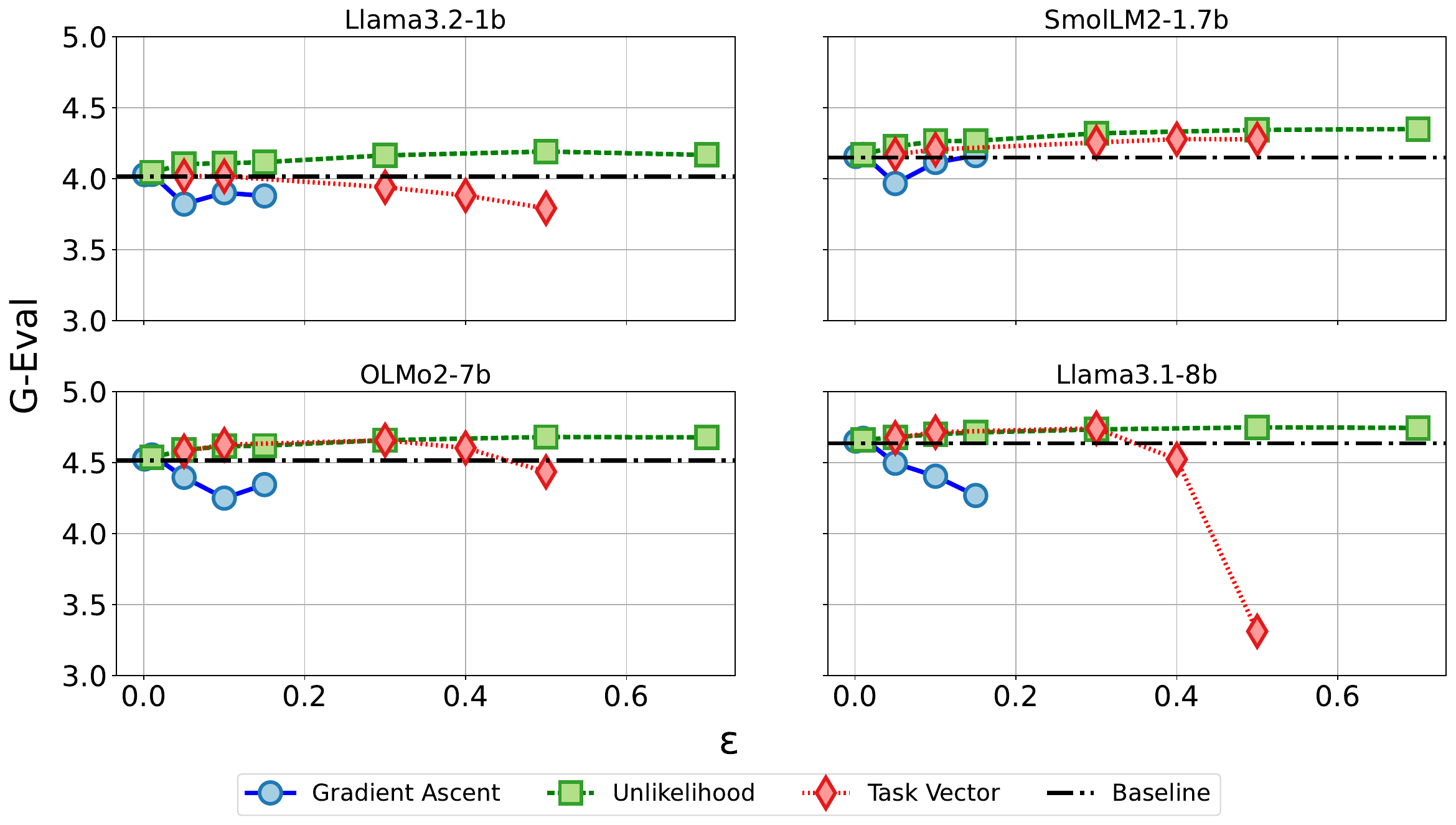}
    \caption{Average G-Eval on all three datasets with models trained using different $\epsilon$.}
    \label{fig:epsilon}
\end{figure*}

In Section~\ref{sec:methods_studied} we introduced a hyperparameter $\epsilon$ as the weight assigned to negative unfaithful samples during model fine-tuning. Figure \ref{fig:epsilon} illustrates the effect of $\epsilon$ on the methods' performance. Gradient ascent slightly improves faithfulness when $\epsilon$ is small (0.01), but performance degrades as $\epsilon$ increases. Task vector negation can generally improve faithfulness, but performance degrades when $\epsilon$ becomes large (except for SmolLM2). Unlikelihood training is the most stable among the three methods, providing consistent improvements even with $\epsilon$ as large as 0.7. 
These results are consistent with the findings in Section~\ref{sec:main-results} that all three methods can improve summarization faithfulness with unlikelihood training and task vector negation being more effective than gradient ascent.

\subsection{Discussion}
\label{sec:discussion}

\paragraph{Effect of $\epsilon$} One prominent phenomenon we observe in Figure~\ref{fig:epsilon} and during training is the three effect of $\epsilon$ across the different methods. Gradient ascent only seems to perform well when $\epsilon$ is very small (i.e. 0.01). Increasing the weight of $\epsilon$ is detrimental to the resulting model's performance to the point of destroying the model's ability to generate coherent text. Task vector negation can tolerate moderate $\epsilon$ values and unlikelihood training is the most robust across different values of $\epsilon$. 
We suspect the tolerance to $\epsilon$ is an important factor that determines a method's ability to leverage negative/unfaithful span-level information to improve faithfulness. This can help explain why gradient ascent only slightly improve faithfulness while the other two methods observe greater improvements.

\paragraph{The inconsistency of BARTScore} Although BARTScore is a widely used metric for evaluating summarization faithfulness, in our experiments, it performs inconsistently and does not correlate well with the other metrics used in this evaluation. For example, Table \ref{tab:baseline_model_results} \& \ref{tab:samsum_result} shows that most finetuned models are lower in BARTScore than the baseline, and gradient ascent has higher BARTScores than unlikelihood training and task vector negation. These opposite conclusions from BARTScore make us suspect the negative token probability-minimizing objective of our training methods results in lower BARTScores, since BARTScore is essentially the average token probability. 
This finding challenges the premise of BARTScore: positive correlation between token probabilities and faithfulness. Thus we include BARTScore for reference but do not draw conclusions from it.

\vspace{-2mm}
\section{Conclusion}

In this study, we propose leveraging span-level hallucination annotations to improve the faithfulness of LLM-generated summaries. To this end, we construct a dataset containing organically generated summaries labeled at the span level and evaluate three fine-tuning methods—\textit{gradient ascent}, \textit{unlikelihood training}, and \textit{task vector negation}—that leverage unfaithful summary spans. Our results show that all three approaches improve summarization faithfulness, with unlikelihood training and task vector negation being more effective. Additionally, we analyze the impact of varying the weight of negative samples ($\epsilon$) and find that unlikelihood training remains the most stable across a wide range of values. These findings highlight the effectiveness of span-level annotations in mitigating hallucinations and offer insights into robust fine-tuning strategies for improving LLM faithfulness.

\vspace{-2mm}
\section{Limitations}
This work investigates the effectiveness of span-level faithfulness annotations in combination with fine-tuning methods that leverage the span-level information to reduce hallucinations in LLM-generated summaries. 
While our primary focus is on leveraging these annotations to improve faithfulness, the reliability of GPT-4o-based span annotation remains an open question. Although not a core contribution of this work, a more rigorous study of the annotation method is necessary to assess its accuracy and consistency. We acknowledge this limitation and intend to explore a more comprehensive validation in future work.


Moreover, \citet{cao-wang-2021-cliff} uses summary-level faithfulness information to improve summarization faithfulness, finding contrastive learning to be the most effective approach. Our work, being more fine-grained, focuses on leveraging span-level information. However, to perform contrastive learning on span (token)-level, requires a corresponding positive token for each negative token, which our current dataset does not provide. As a result, we are unable to include contrastive learning in this study.

Lastly, while our work explores fine-tuning techniques for improving faithfulness, we do not include comparisons with recent alignment methods such as Direct Preference Optimization \citep[DPO][]{NEURIPS2023_a85b405e}. We recognize the growing adoption of these techniques for aligning LLM behavior and plan to investigate their effectiveness in mitigating hallucinations in future work.





\bibliography{colm2025_conference}
\bibliographystyle{colm2025_conference}

\appendix

\section{LLM prompts used in experimentation}
\label{app:prompts}
\vspace{5mm}

\subsection{Prompt for LLM to generate summary - CNNDM}
\label{app:cnn-prompt}
\begin{ttfamily}
\textbf{System prompt:} You are an accurate summarizer that always writes concise summaries of text that is as consistent with the source text as possible. You will be given a piece of news article in the user prompt, and you need to reply with a concise summary of the article as a response.
\\ \\
\noindent \textbf{User prompt:} \{source\_doc\}
\end{ttfamily}
\vspace{10mm}

\subsection{Prompt for LLM to generate summary - SAMSum}
\label{app:samsum-prompt}
\begin{ttfamily}
\textbf{System prompt:} You are an accurate summarizer that always writes concise summaries of text that is as consistent with the source text as possible. You will be given a dialogue conversation in the user prompt, and you need to reply with a concise summary of the conversation in 2 sentences.
\\ \\
\noindent \textbf{User prompt:} \{source\_doc\}
\end{ttfamily}
\vspace{10mm}

\subsection{Prompt for LLM to generate summary - XSum}
\label{app:samsum-prompt}
\begin{ttfamily}
\textbf{System prompt:} You are an accurate summarizer that always writes concise summaries of text that is as consistent with the source text as possible. You will be given a piece of news article in the user prompt, and you need to reply with a concise summary of the article as a response.
\\ \\
\noindent \textbf{User prompt:} \{source\_doc\}
\end{ttfamily}
\vspace{5mm}

\subsection{Prompt for GPT-4o to label spans}
\label{app:gpt-prompt}
\begin{ttfamily}
\textbf{System prompt:} 
\begin{Verbatim}
Your input fields are:
1. `source` (str): The source document.
2. `summary` (str): The summary of the source document.

Your output fields are:
1. `reasoning` (str)
2. `hallucinated_spans` (list[str]): Spans of the summary text that are 
hallucinated i.e. unfaithful to the source (each span must be a substring of the
 summary text).

All interactions will be structured in the following way, with the appropriate 
values filled in.

[[ ## source ## ]]
{source}

[[ ## summary ## ]]
{summary}

[[ ## reasoning ## ]]
{reasoning}

[[ ## hallucinated_spans ## ]]
{hallucinated_spans}        # note: the value you produce must be parseable 
according to the following JSON schema: {"type": "array", "items": {"type": 
"string"}}

[[ ## completed ## ]]

In adhering to this structure, your objective is: 
    Analyze the provided source text and its summary to identify any portions of
 the summary that do not accurately reflect the source content. Provide a 
detailed reasoning for your analysis and list any hallucinated spans—these are 
sections of the summary text that introduce information not present in the 
source or misrepresent the source details. Ensure that your reasoning aligns 
with the source content to justify the identification of hallucinated spans, if 
any.
\end{Verbatim}

\textbf{Demo user prompt 1:}
\begin{verbatim}
This is an example of the task.

[[ ## source ## ]]
Lenny: Babe, can you help me with something?
Bob: Sure, what's up?
Lenny: Which one should I pick?
Bob: Send me photos
Lenny: <picture>
Lenny: <picture>
Lenny: <picture>
Bob: I like the first ones best
Lenny: But I already have purple trousers. Does it make sense to have two pairs?
Bob: I have four black pairs :D :D
Lenny: yeah, but shouldn't I pick a different color?
Bob: what matters is what you'll give you the most outfit options
Lenny: So I guess I'll buy the first or the third pair then
Bob: Pick the best quality then
Lenny: ur right, thx
Bob: no prob :)

[[ ## summary ## ]]
Lenny asked Bob for help picking between three pairs of trousers, and Bob 
recommended choosing the best quality option. Lenny was unsure about picking two
 pairs of the same color and Bob agreed, suggesting that having more options is 
important.

Respond with the corresponding output fields, starting with the field `[[ ## 
reasoning ## ]]`, then `[[ ## hallucinated_spans ## ]]` (must be formatted as a 
valid Python list[str], and the list of spans must follow original order that 
they appear in the summary), and then ending with the marker for `[[ ## 
completed ## ]]`.
\end{verbatim}

\textbf{Demo model response 1:}
\begin{verbatim}
[[ ## reasoning ## ]]
Lenny was unsure about picking two pairs of the same color, but Bob did not 
agree with Lenny's reasoning. Instead, Bob suggested that having more options is
 important.

[[ ## hallucinated_spans ## ]]
["Bob agreed"]

[[ ## completed ## ]]
\end{verbatim}

\textbf{Demo user prompt 2:}
\begin{verbatim}
This is an example of the task.

[[ ## source ## ]]
ARLINGTON, Va. (AP) — Salvage crews have recovered an engine, large sections of 
fuselage, and are working to retrieve a wing from the wreckage of a commercial 
airliner involved in last week’s midair collision near Washington’s Reagan 
National Airport, officials said Monday.

More human remains have been recovered from the Potomac River, bringing the 
number of identified victims to 55 of the 67 aboard the two aircraft. The 
collision between an American Airlines jet and an Army Black Hawk helicopter was
 the deadliest U.S. air disaster since 2001.

Authorities say the operation to remove the plane will take several days, after 
which the helicopter wreckage will be retrieved. More than 300 responders are 
participating in the recovery, with two Navy barges assisting in lifting heavy 
debris. Divers and salvage workers are adhering to strict protocols, pausing 
recovery efforts when human remains are found.

The American Airlines jet, carrying 64 people, was en route from Wichita, 
Kansas, preparing to land when it collided with the Army helicopter, which had 
three personnel aboard on a training mission. There were no survivors. 
Passengers included figure skaters returning from the 2025 U.S. Figure Skating 
Championships and a group of hunters.

Family members of victims were escorted by police to the Potomac River on Sunday
 to pay their respects. Officials hope to recover the jet’s cockpit on Tuesday, 
with parts of both aircraft transported to a hangar for investigation.

Federal investigators are analyzing conflicting altitude data from the 
collision. The jet’s flight recorder showed an altitude of 325 feet (99 meters),
 while control tower data placed the Black Hawk at 200 feet (61 meters), the 
maximum permitted altitude for helicopters in the area. About a second before 
impact, the jet’s pitch changed, but investigators have not determined whether 
it was an evasive maneuver.

Authorities warn against premature speculation about the cause of the crash or 
why the helicopter may have been above its altitude limit. “There are all kinds 
of reasons for altitude deviations—something as simple as a flock of birds or an
 obstacle,” said Col. Mark Ott, deputy director of Army aviation.

The National Transportation Safety Board (NTSB) is working to reconcile the 
conflicting altitude readings using data from the helicopter’s black box and 
refining the tower’s data. All five air traffic controllers on duty at Reagan 
Airport during the collision have been interviewed. A preliminary report is 
expected within 30 days, though a full investigation may take over a year.

[[ ## summary ## ]]
A midair collision near Washington’s Reagan National Airport between an American
 Airlines jet and an Army Black Hawk helicopter resulted in 76 fatalities, 
making it the deadliest U.S. air disaster since 2001. Salvage crews have 
recovered major wreckage, and the entire wreckage will be recovered within a 
day. More human remains have been found, with 55 victims identified. 
Investigators are analyzing conflicting altitude data, as the jet’s flight 
recorder showed 325 feet while control tower data placed the helicopter at 200 
feet. The flight data recorder also indicated the jet made an evasive maneuver 
right before the crash. Authorities caution against speculation as the National 
Transportation Safety Board (NTSB) investigates, with a preliminary report 
expected in 30 days.

Respond with the corresponding output fields, starting with the field `[[ ## 
reasoning ## ]]`, then `[[ ## hallucinated_spans ## ]]` (must be formatted as a 
valid Python list[str], and the list of spans must follow original order that 
they appear in the summary), and then ending with the marker for `[[ ## 
completed ## ]]`.
\end{verbatim}

\textbf{Demo model response 2:}
\begin{Verbatim}
[[ ## reasoning ## ]]
1. There were 67 people in total on board the two aircrafts, resulting in 67 
fatalities. Not 76.
2. Though the engine and fuselage were recovered, and a wing is being retrieved,
 the claim that the entire wreckage will be recovered within a day is incorrect.
3. The flight data recorder indicated the jet's pitch changed a second before 
impact, but the investigators have not determined whether it was an evasive 
maneuver.

[[ ## hallucinated_spans ## ]]
["76", "the entire wreckage will be recovered within a day", "the jet made an 
evasive maneuver"]

[[ ## completed ## ]]
\end{Verbatim}

\textbf{User prompt:}
\begin{Verbatim}
[[ ## source ## ]]
{source}

[[ ## summary ## ]]
{summary}

Respond with the corresponding output fields, starting with the field `[[ ## 
reasoning ## ]]`, then `[[ ## hallucinated_spans ## ]]` (must be formatted as a 
valid Python list[str], and the list of spans must follow original order that 
they appear in the summary), and then ending with the marker for `[[ ## 
completed ## ]]`.
\end{Verbatim}

\end{ttfamily}

\section{More results}
\label{app:more_results}
Here are more results on CNNDM and XSum datasets all showing similar trends.

\begin{table}[H]
\centering
\begin{minipage}{0.48\linewidth}
    \centering
    \begin{tabular}{@{}lccc@{}}
        \toprule
        \textbf{Model} & \textbf{GE} & \textbf{AS} & \textbf{BS} \\
        \midrule
        Llama3.2-1b & 4.139 & 0.851 & -1.857 \\
        SmolLM2-1.7b & 4.292 & 0.866 & -1.878 \\
        OLMo2-7b & 4.641 & 0.892 & -1.917 \\
        Llama3.1-8b & 4.733 & 0.900 & -1.906 \\
        \bottomrule
    \end{tabular}
    \caption{SFT baseline results of four models on CNNDM.}
    \label{tab:cnn_baseline}
\end{minipage}
\hfill
\begin{minipage}{0.48\linewidth}
    \centering
    \begin{tabular}{@{}lccc@{}}
        \toprule
        \textbf{Model} & \textbf{GE} & \textbf{AS} & \textbf{BS} \\
        \midrule
        Llama3.2-1b & 3.875 & 0.855 & -1.772 \\
        SmolLM2-1.7b & 3.993 & 0.864 & -1.778 \\
        OLMo2-7b & 4.379 & 0.893 & -1.849 \\
        Llama3.1-8b & 4.585 & 0.906 & -1.836 \\
        \bottomrule
\end{tabular}
    \caption{SFT baseline results of four models on XSum.}
    \label{tab:xsum_baseline}
\end{minipage}
\end{table}

\begin{table}[H]
\resizebox{\textwidth}{!}{%
\begin{tabular}{lp{2.5cm}cccccccccc}
\toprule
\multicolumn{1}{c}{} & \multirow{2}{*}{\raggedright Model}
& \multicolumn{3}{c}{Gradient Ascent}
& \multicolumn{3}{c}{Unlikelihood}  
& \multicolumn{3}{c}{Task Vector} \\ 
\cmidrule(l){3-5} 
\cmidrule(l){6-8}
\cmidrule(l){9-11}
\multicolumn{1}{c}{} & \multicolumn{1}{c}{} & GE & AS & BS & GE & AS & BS & GE & AS & BS\\
\midrule
& \raggedright Llama3.2-1b  
  & 4.171\(\uparrow\) & 0.860\(\uparrow\) & -1.869  
  & \textbf{4.306}\(\uparrow\) & \textbf{0.885}\(\uparrow\) & -1.817\(\uparrow\)  
  & 4.084 & 0.847 & -1.923  \\

& \raggedright SmolLM2-1.7b  
  & 4.309\(\uparrow\) & 0.869\(\uparrow\) & -1.872  
  & \textbf{4.455}\(\uparrow\) & \textbf{0.894}\(\uparrow\) & -1.874\(\uparrow\)  
  & 4.443\(\uparrow\) & 0.880\(\uparrow\) & -1.934  \\

& \raggedright OLMo2-7b  
  & 4.693\(\uparrow\) & 0.898\(\uparrow\) & -1.906\(\uparrow\)  
  & \textbf{4.767}\(\uparrow\) & 0.918\(\uparrow\) & -1.852\(\uparrow\)  
  & 4.709\(\uparrow\) & \textbf{0.927}\(\uparrow\) & -1.971  \\

& \raggedright Llama3.1-8b  
  & 4.737\(\uparrow\) & 0.901\(\uparrow\) & -1.907  
  & \textbf{4.796}\(\uparrow\) & 0.915\(\uparrow\) & -1.875\(\uparrow\)  
  & 4.785\(\uparrow\) & \textbf{0.924}\(\uparrow\) & -1.930  \\

\bottomrule
\end{tabular}
}
\caption{Results on CNNDM for three studied methods on four models according to G-Eval (GE), AlignScore (AS), and BARTScore (BS). Gradient ascent $\epsilon = 0.01$, unlikelihood $\epsilon = 0.5$, and task vector $\epsilon = 0.3$. The best result (according to G-Eval and AlignScore) for each model is \textbf{in-bold}. Scores higher than the baseline are marked with \(\uparrow\).}
\label{tab:cnn_result}
\end{table}

\begin{table}[h!]
\resizebox{\textwidth}{!}{%
\begin{tabular}{lp{2.5cm}cccccccccc}
\toprule
\multicolumn{1}{c}{} & \multirow{2}{*}{\raggedright Model}
& \multicolumn{3}{c}{Gradient Ascent}
& \multicolumn{3}{c}{Unlikelihood}  
& \multicolumn{3}{c}{Task Vector} \\ 
\cmidrule(l){3-5} 
\cmidrule(l){6-8}
\cmidrule(l){9-11}
\multicolumn{1}{c}{} & \multicolumn{1}{c}{} & GE & AS & BS & GE & AS & BS & GE & AS & BS\\
\midrule
& \raggedright Llama3.2-1b
  & 3.838 & 0.858\(\uparrow\) & -1.767\(\uparrow\)  
  & \textbf{4.100}\(\uparrow\) & \textbf{0.895}\(\uparrow\) & -1.761\(\uparrow\)  
  & 3.741 & 0.841 & -1.824\(\uparrow\)  \\

& \raggedright SmolLM2-1.7b
  & 4.031\(\uparrow\) & 0.868\(\uparrow\) & -1.780  
  & \textbf{4.237}\(\uparrow\) & \textbf{0.898}\(\uparrow\) & -1.803  
  & 4.089\(\uparrow\) & 0.885\(\uparrow\) & -1.798  \\

& \raggedright OLMo2-7b
  & 4.424\(\uparrow\) & 0.906\(\uparrow\) & -1.817\(\uparrow\)  
  & \textbf{4.601}\(\uparrow\) & \textbf{0.921}\(\uparrow\) & -1.794\(\uparrow\)  
  & 4.545\(\uparrow\) & 0.909\(\uparrow\) & -1.950  \\

& \raggedright Llama3.1-8b 
  & 4.632\(\uparrow\) & 0.909\(\uparrow\) & -1.836  
  & \textbf{4.702}\(\uparrow\) & 0.919\(\uparrow\) & -1.808\(\uparrow\)  
  & 4.686\(\uparrow\) & \textbf{0.935}\(\uparrow\) & -1.875  \\

\bottomrule
\end{tabular}
}
\caption{Results on XSum for three studied methods on four models according to G-Eval (GE), AlignScore (AS), and BARTScore (BS). Gradient ascent $\epsilon = 0.01$, unlikelihood $\epsilon = 0.5$, and task vector $\epsilon = 0.3$. The best result (according to G-Eval and AlignScore) for each model is \textbf{in-bold}. Scores higher than the baseline are marked with \(\uparrow\).}
\label{tab:xsum_result}
\end{table}

\end{document}